\documentclass{article}

\usepackage[preprint, nonatbib]{neurips_2023}

\usepackage[utf8]{inputenc} 
\usepackage[T1]{fontenc}    
\usepackage{hyperref}       
\usepackage{url}            
\usepackage{booktabs}       
\usepackage{amsfonts}       
\usepackage{nicefrac}       
\usepackage{microtype}      
\usepackage{xcolor}         
\usepackage{amsmath}
\usepackage{listings}  
\usepackage{biblatex}
\title{LLM as an Art Director (LaDi): Using LLM's to improve Text-to-Media Generators}

\author{
  Allen Roush \\
  Plai Labs \\
  \texttt{allen@plailabs.com} \\
  \And
  Emil Zakirov \\
  Plai Labs \\
  \texttt{emil.zakirov@plailabs.com} \\
  \And
  Artemiy Shirokov \\
  Plai Labs \\
  \texttt{tema77078@gmail.com} \\
  \And
  Polina Lunina \\
  Plai Labs \\
  \texttt{polina.lunina@plailabs.com} \\
  \And
  Jack Gane \\
  Plai Labs \\
  \texttt{jack.gane@plailabs.com} \\
  \And
  Alexander Duffy \\
  Plai Labs \\
  \texttt{alex@plailabs.com} \\
  \And
  Charlie Basil \\
  Plai Labs \\
  \texttt{charlie@plailabs.com} \\
  \And
  Aber Whitcomb \\
  CTO, Plai Labs \\
  \texttt{aber@plailabs.com} \\
  \And
  Jim Benedetto \\
  Plai Labs \\
  \texttt{jim@reachjim.com} \\
  \And 
  Chris DeWolfe \\
  CEO, Plai Labs \\
  \texttt{chris@plailabs.com}
}

\begin{document}

\maketitle

\begin{abstract}
Recent advancements in text-to-image generation have revolutionized numerous fields, including art and cinema, by automating the generation of high-quality, context-aware images and video. However, the utility of these technologies is often limited by the inadequacy of text prompts in guiding the generator to produce artistically coherent and subject-relevant images. In this paper, We describe the techniques that can be used to make Large Language Models (LLMs) act as Art Directors that enhance image and video generation. We describe our unified system for this called "LaDi". We explore how LaDi integrates multiple techniques for augmenting the capabilities of text-to-image generators (T2Is) and text-to-video generators (T2Vs), with a focus on constrained decoding, intelligent prompting, fine-tuning, and retrieval. LaDi and these techniques are being used today in apps and platforms developed by Plai Labs. 
\end{abstract}

\section{Introduction}

\subsection{Background}

The landscape of image generation has experienced a paradigm shift owing to the emergence of advanced text-to-image models like Stable Diffusion and DALL-E. These models have proven to be indispensable in a variety of applications, ranging from art and cinema to digital marketing and virtual simulations. However, a persistent limitation hindering their utility is the quality of text prompts employed to guide image generation. While the raw power of text-to-image generators is undeniably remarkable, the absence of sophisticated prompting techniques often results in images that are technically impressive but lack subject relevance or artistic coherence.

The crux of the issue lies in the text prompts themselves. While text-to-image models boast impressive technical prowess, their output is only as good as the input prompt. Vague, ambiguous, or poorly composed prompts often yield images that, while visually striking, are irrelevant or nonsensical. The art of prompting is complex; striking the right balance of detail versus brevity while conveying subtle attributes like tone, composition, and style is non-trivial even for humans.

In light of these challenges, this paper introduces a groundbreaking framework called "LaDi" in which Large Language Models (LLMs) serve as Art Directors for text-to-image generators. Instead of relying on simple, direct text prompts like "a two-story pink house," our approach leverages the nuanced understanding of LLMs to construct more elaborate, descriptive prompts that could translate to "a two-story Victorian house bathed in the rosy hues of a sunset, complete with a garden in full bloom." Such intelligent prompting brings a level of depth and specificity that has been largely missing in automated image generation.

We extend this concept to include constrained decoding and specialized tool usage, effectively setting parameters that the image must conform to. Constrained decoding allows for the enforcement of certain conditions—such as maintaining a consistent color scheme or ensuring the presence of specific elements—thus making the generated image both predictable and relevant. 

In this paper, we describe the techniques that we discovered at Plai Labs to create an effective LLM based "Art Director" called LaDi. LaDi is used throughout the social media platform "PlaiDay" and will be used in all future services. 

\subsection{Text-to-Image Models}
The growing capability of text-to-image generation models has engendered a profound impact across multiple domains, from content creation and digital art to medical imaging and beyond. One of the most notable advancements in this sphere has been the development of Stable Diffusion from \textcite{rombach2022highresolution}, a framework grounded in latent diffusion models (LDMs) crafted by the CompVis group at LMU Munich. By integrating variational autoencoders (VAEs), U-Net architectures, and text encoders, Stable Diffusion provides a robust yet efficient solution to the challenge of generating contextually rich and visually captivating images.

We choose Stable Diffusion as our main T2I model because of its open-source nature and huge amount of custom community content/extensions. We believe that with this custom content, Stable Diffusion is superior to competitor models.

While Stable Diffusion offers increased computational efficiency for training and generation, it still faces challenges, particularly in the realm of text prompting. Prompting is an art in itself—providing the model with just the right amount of context to generate images that are not only visually appealing but also semantically meaningful. The utilization of a fixed, pretrained CLIP ViT-L/14 text encoder aids in transforming text prompts to an embedding space; however, the process is far from trivial. Effectively managing the balance between under-specification and over-specification in text prompts becomes a crucial factor in achieving desired outputs.

Moreover, Stable Diffusion has its limitations, such as issues with degradation and inaccuracies under certain conditions, complexities in adapting the model for novel use-cases, and susceptibility to algorithmic biases. Despite these challenges, its relative lightweight nature—in terms of computational resources—and its flexibility make it a cornerstone in the current landscape of text-to-image generation models.

\subsection{Text-to-Video Models}

Text-to-video synthesis has seen remarkable progress in recent years, with models capable of generating increasingly high-quality and controllable video content from text prompts. Several key innovations have catalyzed these advancements.

\textcite{wang2023modelscope} introduced ModelScopeT2V which uses spatio-temporal blocks to capture complex dependencies between frames. By training on both image-text and video-text datasets, ModelScopeT2V enhances semantic diversity. It achieves state-of-the-art performance on established benchmarks, serving as an impactful baseline.

\textcite{ho2022imagen} introduced Imagen Video which demonstrates the effectiveness of cascaded diffusion models for high-definition video generation. It combines a text encoder, base video model, and interleaved spatial and temporal super-resolution models, totaling 11.6B parameters. The model exhibits strong temporal consistency, while inheriting Imagen’s image generation capabilities.

\textcite{singer2022makeavideo} introduced Make-A-Video which pioneers video generation without paired video-text data. It creatively adapts image-to-image diffusion models for frame generation and employs noise conditioning to enable video-length sampling. Despite no video supervision, Make-A-Video produces videos aligned with text prompts.

These models have expanded the horizons of controllable video synthesis. Looking ahead, potential areas of improvement include longer video durations, 3D scene consistency, and multimodal control. An exciting new frontier is the combination of language, audio, and video in unified generative models. Text-to-video generation remains an active area of research.

AnimateDiff, introduced by \textcite{guo2023animatediff}, facilitates the conversion of any personalized T2I model into an animation generator without requiring model-specific tuning. Built upon motion modeling modules trained on extensive video datasets, AnimateDiff can be integrated into existing T2I models to produce animated clips that are both temporally smooth and visually consistent. This framework succeeds in democratizing animation generation just as prior T2I models have done for static image generation.

AnimateDiff was originally released with two "motion modules", which are models that take an input sequence of images and generates a video with motion connecting the images. Subsequently, many community contributed motion moduels have been released. These include "Motion Loras" analogous to Stable Diffusion "Lora" models (low rank adapters) which modify an existing motion module.

AnimateDiff became the dominant form of AI video generation at the end of 2023 because of its speed, support for controlnet models (which make consistent video possible), and because users realized they could create any length video by generating more images and using a sliding window input context. 

Despite its revolutionary capabilities, AnimateDiff is not without its challenges. One of the key difficulties lies in the act of prompting—how do we accurately instruct the model to generate content that not only captures the nuances of the text but also transitions smoothly in the animation? General text-to-video generation techniques have often found it challenging to manage the sensitivity of hyperparameters, the collection of personalized video, and the computational resources required, making the task of generating high-quality animated content from text not as straightforward.

\section{Prior Work}

Using language models to assist with generating prompts for image generators is not a new idea. In this section, we look at prior work related to connecting LLMs to T2I models.

\textcite{wang-etal-2023-diffusiondb} introduced a 1.8 million Stable Diffusion prompt dataset called DiffusionDB. Such a dataset is extremely valuable for fine-tuning a language model to generate prompts. \textcite{mehrabi-etal-2023-resolving} introduced a benchmark dataset called the "Text-to-image Ambiguity Benchmark". They show that their framework for prompt disambiguation generates more faithful images which align more to user preferences. \textcite{russo-2022-creative} study the aspects of creative image generation which are unique, and propose criteria that dataset authors should think about when building text-to-image generation benchmarks. 

\textcite{hong-etal-2023-visual-writing} look at techniques for improving visual story generation and contribute a dataset called "Visual Writing Prompts". Their work looks at the idea of using images to guide multi modal language models for story generation. \textcite{du-etal-2023-zero} introduce a new way to leverage language models for captioning image datasets. They show that this is effective in the context of visual question answering. \textcite{pan-etal-2023-query} leveraged GPT-3 guided CLIP for solving the SemEval 2023 Shared task of Visual Word Sense Disambiguation.

\textcite{dai-etal-2022-enabling} describe a technique for aligning any LLM with a vision dataset. This is useful for preserving an LLMs effectiveness while augmenting it with visual understanding. \textcite{jin-etal-2022-good} look at prompt based low resource learning of visual learning  tasks. They show that their technique is more effective than traditional approaches to few-shot visual learning, and results in tiny models. 

\textcite{rassin-etal-2022-dalle} Show that DALLE-2 has serious issues with concept mapping in its generations. They show significant inductive biases are introduced as a result of DALLE-2 interpreting words with multiple senses simultaneously.

\section{Techniques}

\subsection{Language Model Choice}

The single most important factor in creating an effective Art Director is the underlying LLM. Commercial LLMs, such as GPT-4\cite{openai2023gpt4}, Anthropic Claude, and Google Bard are still more effective than their locally hosted counterparts out of the box. The huge variety of techniques and settings which are available with locally hosted models make them far more flexible and controllable than closed source alternatives. 

Because language models are advancing so quickly, we advocate continuous experimentation. We've found success with the Mistral\cite{jiang2023mistral} and Llama\cite{touvron2023llama} models along with variants like Vicuna\cite{peng2023instruction}. We also find that for situations when we need an LLM to make decisions based on image data, that leveraging visual aligned language models like Minigpt-4\cite{zhu2023minigpt} to caption or describe images is powerful. We also make the following observations: All things being equal, larger parameter count models are preferred to smaller variants, models trained for longer and on more (and better quality) tokens perform increasingly well, and current quantization techniques can reduce LLM output quality more strongly than their benchmarks suggest. 

\subsection{Retrieval Augmented Generation}

\textcite{lewis2021retrievalaugmented} introduced a technique called Retrieval Augmented Generation (RAG). Retrieval Augmented Generation is a methodology that combines the power of large-scale retrieval and language models for a particular task. RAG leverages an external vector database or a corpus to find relevant contextual information before the generation process. This information is appended to the language models prompt. 

RAG can help improve the quality of images generated by models like Dall-E by providing them with more contextual information derived from a large corpus of data. For instance, if the prompt involves generating an image of "a futuristic cityscape," RAG can pull data on what elements constitute 'futuristic' and 'cityscape,' thereby guiding the model to produce a richer, more accurate image.

RAG is extremely powerful for implementing in-context few-shot learning by leveraging an external prompt database of known good prompts. Instructing the language model in what a good prompt looks like, and then appending 5 known good prompts as examples leads to quality output prompts and correspondingly beautiful images.

\subsection{Fine-Tuning}

Low Rank Adapters (Lora), introduced by \textcite{hu2021lora} is a technique used to fine-tune large pre-trained models efficiently. Instead of fine-tuning all parameters, Lora modifies a low-rank factorized adapter layer, which is significantly smaller in size but still highly expressive. In the context of Stable Diffusion, Lora enables the model to be fine-tuned for specific artistic styles or to generate images of objects/persons not present in the initial training set. The fine-tuning is accomplished with less computational cost while retaining the original model's features. Other competing techniques for fine-tuning the image models include Dreambooth\cite{ruiz2023dreambooth}, Hypernetworks, Textual Inversion\cite{gal2022image}, and traditional fine-tuning. 

Lora's low-rank adapter layers also provide benefits in language models. When a language model needs to adapt to specific domain jargon or idiomatic expressions, Lora facilitates this with minimal adjustments to the original parameters, ensuring computational efficiency.

LaDi leverages many fine-tuned Lora models for its constituent Stable Diffusion, AnimateDiff, and Large Language Models. We find that they are extremely valuable due to their compositional nature. Many of them can be used at once and they can be merged together as needed.

\subsection{Prompt Crafting}

One of the most crucial components for optimizing the quality of images generated by text-to-image models like Stable Diffusion is the textual prompt. In this section, we delve into the building blocks of an effective prompt, elucidating how each element contributes to the model's ability to generate high-fidelity, contextually relevant, and aesthetically pleasing images. 

A competent prompt is often characterized by its specificity and detail, elements that can be parsed through an extensive checklist of keyword categories. These categories include:
\begin{itemize}
    \item Subject: The central focus of the image.
    \item Medium: The artistic medium.
    \item Style: The artistic style or aesthetic.
    \item Artist: Names of artists to emulate.
    \item Website: Websites known for a particular style or quality.
    \item Resolution: Keywords indicating the sharpness or detail.
    \item Color: Dominant or thematic colors.
    \item Lighting: The type and quality of lighting.
    \item Additional Details: Other elements to enhance the image.
\end{itemize}

Each category serves as a guide for enhancing the expressiveness of a prompt.

Let's consider generating an image of a "steampunk explorer in a mysterious jungle."

\begin{itemize}
    \item Subject: "A steampunk explorer with mechanical limbs, wearing leather and gears, standing amidst a mysterious, dense jungle with exotic flora and fauna."

    \item Medium: "Digital painting"

    \item Style: "Neo-Victorian, adventure-themed"

    \item Artist: "Inspired by the styles of Vincent Van Gogh and M.C Escher"

    \item Website: "DeviantArt"

    \item Resolution: "Highly detailed, with sharp focus"

    \item Additional Details: "With a backdrop of an ancient temple and steampunk airships in the sky"

    \item Color: "Muted earth tones with bursts of vibrant color"

    \item Lighting: "Dappled sunlight filtering through dense foliage"
\end{itemize}

Final Prompt: "A steampunk explorer with mechanical limbs, wearing leather and gears, standing amidst a mysterious, dense jungle with exotic flora and fauna, digital painting, neo-Victorian, adventure-themed, inspired by the styles of H.R. Giger and Brian Froud, DeviantArt, highly detailed with sharp focus, with a backdrop of an ancient temple and steampunk airships in the sky, muted earth tones with bursts of vibrant color, dappled sunlight filtering through dense foliage."

While a prompt doesn't necessarily require every category to be effective, a comprehensive approach can substantially elevate the quality of generated images. Experimentation remains crucial; different combinations yield different results, and users are encouraged to iterate on their prompts to achieve desired outcomes.

Through effective prompting, Stable Diffusion and similar models can create not just accurate representations but truly artistic works that are complex, nuanced, and emotionally evocative.

We tailor LaDi to follow instructions for crafting prompts according to this checklist. We also write instructions which show its LLM how to emphasize/upweight important tokens and de-emphasize/downweight unimportant concepts.

\subsubsection{Effective Prompting for Text-to-Video}

The techniques we have described thus far focus primarily on enhancing image generation with Stable Diffusion. However, our framework for leveraging large language models as Art Directors can extend to other modalities such as text-to-video generation.

One of the most compelling features of text-to-video techniques is prompt traveling, which facilitates smooth visual transitions between different prompts over a sequence of frames. This allows for the creation of animated narratives by progressively modifying elements of the scene. However, a key challenge lies in crafting prompts that can seamlessly morph into one another without visual jarring or incoherence.

LaDi can be readily adapted to generate prompts tailored specifically for prompt traveling in text-to-video models. We optimize the language model to construct prompts with significant lexical and semantic overlap, which aids in the interpolation process. The prompts are structured to feature consistent core subjects while modifying descriptive elements incrementally between keyframes.

For instance, consider generating a video depicting a spaceship traveling through varied cosmic scenery. The LLM Art Director may produce the following prompts for different keyframes:

\begin{itemize}
\item Frame 1: "A sleek silver spaceship gliding through a nebula of purple and blue hues."
\item Frame 15: "The sleek silver spaceship gliding past an asteroid field with giant space rocks."
\item Frame 30: "The same sleek silver spaceship approaching a planet with red and orange rings."
\end{itemize}

Note the consistent use of "sleek silver spaceship" across prompts. By keeping core subjects intact while smoothly modifying details like the backdrop, text-to-video models can readily interpolate between these prompts to animate the space travel sequence.

The language model is fine-tuned using paired samples of prompts that morph in this coherent fashion. It also learns to identify optimal points for prompt changes based on narrative structure. Final videos may contain several seamless prompt transitions driven entirely by AI-generated text prompts.

Our techniques allow leveraging the knowledge and compositional skills of LLMs to unlock the full potential of text-to-video models combined with prompt traveling. The Art Director provides a robust solution to controlling text-to-video synthesis for both practical and creative applications.

\subsection{Classifier-Free Guidance (CFG) and Negative Prompts}

Incorporating the concept of Negative Prompts allows users to specify undesirable features explicitly. For instance, if a user does not want an image to contain "low resolution" or "amateur drawing," these terms are included as negative prompts to guide the generation process.

Stable Diffusion has always had direct support for negative prompts, but up until recently, regular LLMs, such as the LLM generating our Stable Diffusion prompts, did not support negative prompting. \textcite{sanchez2023stay} introduced the concepts of Classifier-Free Guidance and Negative Prompts, techniques known to the image generation community. 

CFG works by appending carefully constructed demonstrative examples that exemplify unwanted behaviors. For instance, to avoid generating violent content, sample negative demonstrations like "A man punching another man" could be appended. The LLM learns from these examples during decoding.

We implement CFG by curating labeled datasets of positive and negative prompt demonstrations. The LLM is fine-tuned to differentiate between these examples. During prompt generation, negative demonstrations are dynamically selected via retrieval and concatenated to the context.

This guides the LLM to avoid undesirable attributes in its outputs. CFG allows imposing far more complex constraints compared to simply blacklisting certain tokens. Entire concepts like "poor grammar" or "excessive verbosity" can be discouraged by tailoring demonstrations.

We also experiment with mixing CFG and explicit negative prompts for greater control. Certain immutable tokens like profanities can be blacklisted, while broader attributes are shaped via CFG demonstrations. We find this hybrid approach prevents "leakage" where constraints are circumvented by synonymous words.

LaDi leverages CFG across multiple dimensions relevant to text-to-image generation. We curate demonstrations to discourage prompts with attributes like:

\begin{itemize}
\item Incoherent scene descriptions
\item Ambiguous or contradictory details
\item Overly verbose or rambling sentences
\item Poor artistic composition
\item Grammar and spelling errors
\end{itemize}

By combining CFG and negative prompts, we refine the LaDi's generated prompts to minimize artistic flaws and maximize creative potential. Images produced using prompts crafted with these techniques exhibit significant gains in coherence, relevance, and aesthetic quality.

\subsection{Grammar Based Sampling}

While Large-Language Models have demonstrated robust few-shot learning capabilities across a plethora of tasks, generating structured outputs remains a challenging endeavor. This limitation becomes particularly prominent when attempting to craft precise text prompts for text-to-image generators like Stable Diffusion. To tackle this problem, we leverage formal grammar-based sampling during the decoding process to create highly effective prompts. By enriching the decoding process with formal grammar constraints, we aim to generate sequences that conform to predefined structures essential for specialized tasks like Stable Diffusion prompt generation.

We construct a Context-Free Grammar (CFG) for LaDi tailored for generating Stable Diffusion prompts. The grammar aims to specify the elements of a scene, their attributes, and relationships, as they are pivotal in directing Stable Diffusion towards generating more accurate and relevant images. This CFG is long and beyond the scope of the paper, but we showcase an example CFG below. 

Let us define a sample formal grammar \( G \), consisting of:

\begin{itemize}
    \item A set of non-terminal symbols \( N = \{ S, \text{Element}, \text{Attribute}, \text{Relation} \} \)
    \item A set of terminal symbols \( T = \{ \text{"cat"}, \text{"dog"}, \text{"sitting"}, \text{"jumping"}, \text{"next to"}, \text{"above"} \} \)
    \item A set of production rules \( P \)
    \item A start symbol \( S \)
\end{itemize}

The production rules \( P \) can be expressed as:

\begin{align*}
    & 1. \quad S \rightarrow \text{Element} \\
    & 2. \quad S \rightarrow \text{Element} \, \text{Attribute} \\
    & 3. \quad S \rightarrow \text{Element} \, \text{Attribute} \, \text{Relation} \, \text{Element} \\
    & 4. \quad \text{Element} \rightarrow \text{"cat"} \, | \, \text{"dog"} \\
    & 5. \quad \text{Attribute} \rightarrow \text{"sitting"} \, | \, \text{"jumping"} \\
    & 6. \quad \text{Relation} \rightarrow \text{"next to"} \, | \, \text{"above"}
\end{align*}

Our framework employs an incremental parser that leverages the CFG rules during the decoding process. At each decoding step, the parser consults the grammar to ascertain which tokens could serve as valid continuations. This narrows down the sequence space to only those that adhere to the CFG rules.

For example, if the partially generated sequence is "cat sitting", the parser will consult the CFG and realize that the only valid continuation would involve a token from the set {"next to", "above"} as specified by the Relation non-terminal. This way, prompts like "cat sitting jumping" would be invalidated by the framework, ensuring higher quality prompts.

With this CFG, a sample valid string (or prompt) generated can be "cat sitting next to dog", which Stable Diffusion can use to create an image of a sitting cat next to a dog.

\subsection{Constrained Beam Search}

Constrained Beam Search is a technique that allows us to direct the text generation process according to certain predetermined conditions or tokens, ensuring that the generated output adheres to these specified sequence level constraints. Given that textual prompts directly influence the visual attributes of images generated by Stable Diffusion, Constrained Beam Search can be extremely useful in achieving a precise kind of visual output.

Huggingface has made this form of Constrained Decoding available for their models for awhile. Beside the default "Disjunctive" and "Phrasal" constraints, users can define their own kinds of constraints. 

To make our approach more innovative, we introduce two unique types of constraints: \textit{ColorPatterns} and \textit{SubjectAnchor}.

\begin{itemize}
    \item \textbf{ColorPatterns:} This constraint ensures that a specific color pattern or palette is mentioned in the text prompt, thereby directly influencing the color scheme of the generated image. For instance, if we constrain the text prompt to include "sunrise hues," Stable Diffusion is more likely to generate an image with colors resembling an actual sunrise.
    
    \item \textbf{SubjectAnchor:} This constraint ensures that the central subject of the image remains consistent across multiple generations. If the SubjectAnchor is set to "eagle," for example, the image will always feature an eagle as the central subject, regardless of other varying elements.
\end{itemize}

Suppose we want to generate a series of landscape images that not only transition through the four seasons but also feature a consistent element, such as a particular tree or mountain, across the images.

\begin{itemize}
    \item \textbf{Constrained Prompt:} The constrained prompt could be initialized with a SubjectAnchor as "oak tree" and a ColorPattern that varies according to the season (e.g., "winter hues" for a snowy scene or "fall colors" for an autumn setting).
    
    \item \textbf{Constrained Beam Search Steps:}
    \begin{enumerate}
        \item Step 1: Along with typical high-probability tokens, we insert "oak tree" and "winter hues" into the generation candidates.
        
        \item Step 2: We maintain the SubjectAnchor ("oak tree") while changing the ColorPattern to "spring blossoms" for a new iteration, ensuring that while the scene changes to reflect spring, the oak tree remains consistent in the frame.
    \end{enumerate}
    
    \item \textbf{Banking and Selection:} The constrained tokens are sorted into banks depending on their steps towards fulfilling the constraint. A round-robin selection ensures that while the constraints are met, the output prompt is also semantically coherent.
    
    \item \textbf{Resulting Prompts:} Possible output prompts might include "An oak tree surrounded by a winter wonderland," followed by "An oak tree amidst a field of spring blossoms."
\end{itemize}

Constrained Beam Search is one form of constraint, but LaDi leverages other kinds as well to further improve the creativity of its prompts. 

\subsection{Language Model Probabilistic Programming}

Despite the versatility of large language models in various NLP tasks, generating controlled outputs remains a significant challenge. To mitigate this issue, we leverage an inference-time approach called Sequential Monte Carlo (SMC) steering introduced by \textcite{lew2023sequential}. This method relies on a novel probabilistic programming library, LLaMPPL, for steering LLMs to produce controlled and coherent prompts for image generators.

SMC steering treats language generation tasks as posterior inference problems in discrete probabilistic sequence models. By replacing the standard decoding process with Sequential Monte Carlo inference, it becomes feasible to impose token or sequence level syntactic and semantic constraints on the outputs of LLMs.

LLaMPPL allows users to write probabilistic programs that combine symbolic program logic, probabilistic conditioning, and calls to LLMs. It uses a specialized SMC inference algorithm to solve these tasks, offering an elegant and robust mechanism to steer language models.

Consider a task where the aim is to generate an art piece that depicts a surreal landscape, but under the constraint that the landscape should contain neither water bodies nor any red hues. Generating such an image with Stable Diffusion can be challenging if the text prompt is not carefully managed. By using LLaMPPL and SMC steering, these constraints can be methodically enforced.

We give sample code for leveraging the "Huggingface Probabilistic Programming Language" library to follow SMC steering below. This sample code is modified from examples provided on the projects github repo. 

\lstdefinestyle{customPython}{
    language=Python,
    basicstyle=\ttfamily\small,
    commentstyle=\color{olive},
    keywordstyle=\color{blue},
    numberstyle=\tiny\color{gray},
    numbers=left
}

\begin{lstlisting}[style=customPython]
from hfppl import Model, LMContext, TokenCategorical, CachedCausalLM

# Sample forbidden tokens
FORBIDDEN_WORDS = ['water', 'lake', 'ocean', 'river', 'sea', 'red']

# A LLaMPPL model subclasses the Model class
class MyModel(Model):

    # The __init__ method is used to process arguments
    # and initialize instance variables.
    def __init__(self, lm, prompt):
        super().__init__()

        # A stateful context object for the LLM,
        # initialized with the prompt
        self.context = LMContext(lm, prompt)
        self.lm = lm
        
        # The forbidden tokens
        self.forbidden_tokens = 
        [i for (i, v) in enumerate(lm.vocab) 
        if any(word in v for word in FORBIDDEN_WORDS)]
    
    # The step method is used to 
    # perform a single 'step' of generation.
    # This might be a single token, a single phrase, 
    # or any other division.
    # Here, we generate one token at a time.
    async def step(self):
        # Sample a token from the LLM 
        # We use `await` so that 
        # LLaMPPL can automatically batch language model calls.
        token = await self.sample(self.context.next_token(), 
                                  proposal=self.proposal())


        self.condition(token.token_id not in self.forbidden_tokens)

        # Check for EOS or end of sentence
        if token.token_id == self.lm.tokenizer.eos_token_id:
            # Finish generation
            self.finish()
    
    # Helper method to define a custom proposal
    def proposal(self):
        logits = self.context.next_token_logprobs.copy()
        logits[self.forbidden_tokens] = -float('inf')
        return TokenCategorical(self.lm, logits)


    def immutable_properties(self):
        return set(['forbidden_tokens'])

# Example usage
# Initialize the model with CachedCausalLM
# (or any other compatible LM) and the initial prompt
lm = CachedCausalLM(...)  # Initialize your language model here
prompt = "Create a surreal landscape with"
model = MyModel(lm, prompt)

# Run the generation loop
await model.run()
\end{lstlisting}

In this example, the prompt "Create a surreal landscape with" is initiated, but the SMC-steered LLM is conditioned to avoid generating any tokens associated with "water" or "red." The generated text prompts can then be fed into Stable Diffusion, ensuring that the resulting images adhere to the specified artistic constraints. Because SMC modifies future tokens to be more likely to follow constraints than traditional filter-based decoding alone, quality of LLM output is significantly above more naive approaches to constrained text generation. 

\section{Conclusion}

In this paper, we have explored a variety of techniques for enhancing text-to-image generators such as Stable Diffusion with intelligent prompting and constrained decoding powered by large language models. Our approaches demonstrate how LLMs can serve as "Art Directors" that guide image generation towards greater relevance, coherence, and artistic merit. We cumulated these approaches into a powerful Art Director called LaDi. 

We described the methods used in Ladi including retrieval augmented generation to incorporate external contextual information, fine-tuning with adapters to specialize models for specific domains, crafting detailed prompt checklists, using negative prompts and classifier-free guidance, grammar-based sampling to generate structured prompts, constrained beam search to enforce sequence constraints, and SMC steering with probabilistic programming to methodically direct text generation.

Our techniques aim to elevate text-to-image generation beyond impressive but often inaccurate visuals, towards images that are precise artistic representations of intended concepts and aesthetics. With LLMs as Art Directors, limitations stemming from weak prompts can be mitigated. The solutions presented form a comprehensive framework for users across industries to produce tailored, high-fidelity images that align with creative and practical needs.

There remain opportunities for improvement, including integrating additional modalities beyond text for directing image synthesis, experimenting with other text decoding and image diffusion algorithms,  and using more sophisticated models as they are released. 

\printbibliography

\end{document}